\pdfoutput=1
\documentclass[conference]{IEEEtran}

\usepackage{amsmath}
\usepackage{balance}
\usepackage{booktabs}
\usepackage{cite}
\usepackage{graphicx}
\usepackage{array}
\usepackage{microtype}
\usepackage{multirow}
\usepackage{url}


\newcommand{\system}{Nuni}

\title{Ask Before It Tells: Benchmark-to-Robot Body-Cue Transfer for a Question-First Bedside Robot}

\author{
\IEEEauthorblockN{Dongsik Yoon}
\IEEEauthorblockA{HDC LABS, Republic of Korea\\
kevinds1106@hdc-labs.com}
}

\begin{document}
\maketitle

\begin{abstract}
Body-cue recognition can support assistive robots, but benchmark accuracy does not guarantee reliable behavior under a robot-camera viewpoint. We present \system, a bedside robot prototype that treats a detected distress cue as a reason to ask rather than a reason to alert. We compare two X3D-UGT RGB appearance classifiers, which reach 97.7\% and 94.8\% six-way accuracy on NTU RGB+D, with a pose-centric hybrid pipeline on 28 single-actor scripted clips recorded from the robot camera. The hybrid path achieved 0.71 six-way macro recall, versus 0.25 and 0.29 for the fine-tuned and from-scratch RGB variants. More importantly for interaction, it produced a question-triggering distress cue in 12/16 distress clips and would have prompted unnecessarily in 2/8 normal clips; the RGB variants yielded a question-triggering cue in only 2/16 and 3/16 distress clips. We separately tested the question-first controller through event injection. All 13 state-transition trials passed: valid responses caused stand-down, two unanswered prompts produced one alert, and three boundary conditions were handled correctly. These results are a preliminary technical evaluation, not a user study or medical validation, but they show how interaction policy can limit the consequences of uncertain perception.
\end{abstract}


\section{Introduction}
Socially assistive robots provide aid through interaction rather than physical contact \cite{feilseifer2005sar}. For a robot monitoring a person at home, however, a visual body cue is not equivalent to a verified emergency. Viewpoint, scale, appearance, and pose-estimation errors can shift substantially from benchmark data to a robot camera. A false positive may cause an intrusive escalation, while a silent miss prevents the robot from engaging at all. These outcomes are obscured by a single multiclass accuracy value.

A recurring lesson from deployed vision systems---in image restoration, for example, benchmark-trained super-resolution models fall short on real-world inputs \cite{yoon2025sr}---is that strong benchmark scores do not by themselves survive the shift to deployment conditions. Body-cue recognition on a bedside robot directly inherits this risk.

\system{} builds on three related lines of work. Vision-based fall and distress detection is well studied, but reviews consistently identify real-life robustness, usability, and acceptance as open deployment barriers \cite{igual2013fall,wang2020fallsurvey}, and studies of healthcare robots for older adults likewise report that acceptance depends on how a monitoring system behaves, not only on its accuracy \cite{broadbent2009acceptance,robinson2014role}. Large-scale action-recognition benchmarks adopt cross-subject, cross-view, and cross-setup protocols because viewpoint and context changes degrade recognition \cite{shahroudy2016ntu,liu2020ntu120,kay2017kinetics}; a bedside robot camera is an extreme instance of such cross-context shift for nonverbal body-cue recognition.

Robots can also involve people when autonomous inference is ambiguous. Prior work formalizes the value of asking for help under uncertainty \cite{ren2023askhelp}, while legible robot motion aims to make the robot's intent easy to infer \cite{dragan2013legibility}. Confirmation-before-alert itself is established practice in fall detection: assistive robots detect fallen people \cite{maldonado2019fallen}, multi-stage pipelines dispatch a robot to visually verify a suspected fall before notifying anyone \cite{azghadi2025fall,chen2024wifi}, and voice-based monitors query the user and escalate only on a missing or negative response \cite{tirziu2025fall}. \system{} adopts this \emph{ask-before-tell} policy rather than claiming it as new; the contribution lies in the evidence assembled around it. The policy does not claim to diagnose a condition. It changes what the robot does with a fallible cue, treating the person's own response as part of the sensing loop---a small instance of human--robot cooperative intelligence.

This paper makes three contributions. First, we quantify benchmark-to-robot transfer over a multi-cue distress vocabulary---not a single fall event---by evaluating RGB and pose-centric perception paths, with benchmark-side accuracies for reference, on the same 28 scripted robot-camera clips. Second, we separate exact cue recognition from interaction-relevant trigger outcomes: question triggers, silent misses, unnecessary questions, and waving responses. Third, we validate the question-first finite-state controller with 13 event-injection trials, isolating control reliability from recognition accuracy.

\section{System Design}
\system{} is a bedside prototype built around a commodity consumer pan--tilt camera (TP-Link Tapo C220) on a nightstand; perception consumes the camera's $640\times360$ RTSP substream, and all runtime modules communicate as ROS~2 nodes. An NVIDIA Jetson Orin Nano is the on-device deployment target---our prior work characterizes the RGB backbone on this hardware \cite{yoon2026x3dugt}---while the perception evaluation reported here ran offline on a workstation CPU with ONNX Runtime, over clips recorded from the robot camera.

\subsection{Body-cue perception}
The evaluated vocabulary contains four distress cues---\texttt{clutching\_chest}, \texttt{headache}, \texttt{possible\_collapse}, and \texttt{staggering}---one response cue (\texttt{waving}), and \texttt{normal}. The two appearance baselines are X3D-UGT classifiers with the same six outputs. X3D-UGT is a compact, sub-1M-parameter RGB-only action-recognition network from our prior work \cite{yoon2026x3dugt} that combines an X3D-style hierarchy \cite{feichtenhofer2020x3d} with temporal shift \cite{lin2019tsm} and other efficiency-oriented primitives; on the standard NTU RGB+D 60/120 cross-subject protocols it reaches 95.1\% and 90.9\% top-1. Both six-way variants were trained on NTU RGB+D videos \cite{shahroudy2016ntu,liu2020ntu120} remapped to the evaluated vocabulary (five NTU action classes---hand waving, staggering, falling down, headache, and chest pain---plus a \texttt{normal} class): one was fine-tuned from an NTU-pretrained X3D-UGT checkpoint, and one was trained from scratch for 200 epochs. On a 1,928-clip six-way NTU validation subset (cross-subject) they reach 97.7\% and 94.8\% top-1, respectively; because checkpoints were selected on the same subset, these benchmark values are optimistic. Inputs are 16 uniformly sampled frames, person-cropped with a pose-derived global bounding box and resized to $224\times224$.

The alternative is a \emph{pose-centric hybrid pipeline} that combines four components. YOLO11-Pose \cite{yolo11ultralytics} extracts 17-keypoint COCO-format poses \cite{lin2014coco}; instantaneous pose rules detect response gestures (waving and palm presentation), of which waving is the evaluated response cue; a 48-frame ST-GCN++ component \cite{duan2022pyskl} models chest-, head-, and staggering-related motion; and bounding-box transition geometry supplies a possible-collapse cue. ST-GCN++ follows the graph-based skeleton-recognition line established by ST-GCN \cite{yan2018stgcn} and still advancing \cite{lee2023hdgcn}. Persistence and debounce logic convert frame or window predictions into confirmed tags: keypoints below 0.35 confidence are discarded; chest-, head-, and staggering-related tags require five consecutive confirmations, while gesture and collapse tags require two; and the collapse rule fires only on a recent vertical-to-horizontal transition (bounding-box width/height ratio $\geq 1.35$ within a 4~s window), so a person statically lying in bed does not trigger it. The two perception paths are evaluated separately and are not fused.

\begin{figure*}[t]
    \centering
    \includegraphics[width=0.98\textwidth]{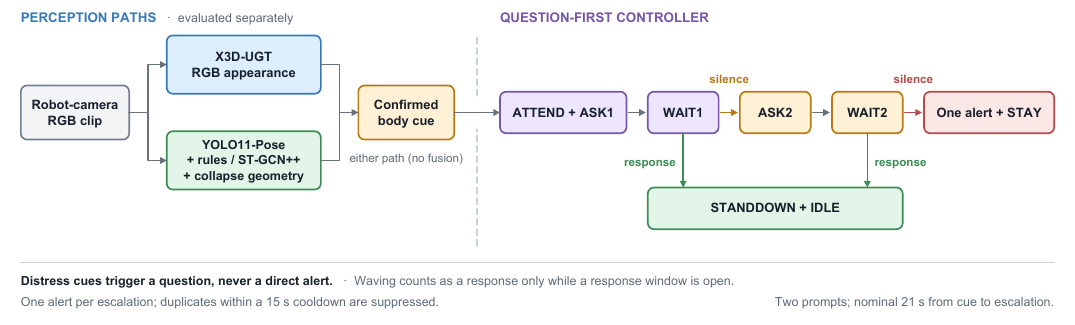}
    \caption{Perception paths and the question-first controller. The RGB and pose-centric paths were evaluated separately and are not fused; either path can produce a confirmed cue. A confirmed distress cue opens a response window instead of directly producing an alert; responses are accepted only during the wait windows (\textsc{Wait1}, \textsc{Wait2}), and both lead to the same stand-down state.}
    \label{fig:system}
\end{figure*}

\subsection{Question-first interaction}
Figure~\ref{fig:system} summarizes the controller. From \textsc{Idle}, or from \textsc{Attend}, in which the robot orients toward the person, any confirmed distress cue enters \textsc{Ask1}. A text prompt is shown for 2.5~s, followed by the 8~s \textsc{Wait1} window. A valid response event, including a waving cue received during a wait state, enters \textsc{Standdown} and returns to \textsc{Idle} after 2~s. Silence causes a second 2.5~s prompt and an 8~s \textsc{Wait2} window. Continued silence publishes one alert and enters \textsc{Stay}. Responses outside \textsc{Wait1} or \textsc{Wait2} are ignored, and a 15~s cooldown suppresses duplicate alerts. The two-prompt schedule bounds the nominal time from a confirmed cue to escalation at $2.5+8+2.5+8=21$~s. Controller states map to a small pan--tilt behavior vocabulary---attend, acknowledge/stand down, and stay (Appendix~\ref{app:design})---but the motions were not physically executed in the reported trials; the present evaluation tests controller state outputs only.

\section{Evaluation}
\subsection{Robot-camera perception protocol}
We recorded 28 trimmed, single-actor scripted clips using the robot camera: four clips for each distress cue (16 total), four waving clips, and eight normal clips containing no distress cue. Each action was held or repeated for roughly eight seconds at 1--2~m from the camera, in frontal and side views. Every perception path processed the same videos in streaming fashion: the pose path applies the persistence rules of Sec.~II-A to sliding 48-frame windows, while the RGB paths were scored more leniently, counting a clip as a hit if any 48-frame span (stride 16) produced the tag, without debounce---so the comparison does not disadvantage the RGB baselines. A clip-level hit required the expected confirmed tag. Six-way macro recall averages per-class recall, with normal counted as correct when no distress tag was confirmed.

We also report interaction-aware measures. \emph{Exact distress} requires the correct one of four distress labels. \emph{Any distress $\rightarrow$ question} counts any distress label, because each opens the same confirmation policy. A \emph{silent miss} contains no confirmed distress label. \emph{Normal $\rightarrow$ question} is an unnecessary prompt on a normal clip, and \emph{waving response} requires a confirmed waving tag. These are technical trigger metrics, not estimates of real-world prevalence or clinical performance.

\begin{table*}[t]
\centering
\caption{Robot-camera perception results on the 28 clips, with each path's six-way NTU benchmark accuracy for reference. Counts show clip-level outcomes. Higher is better for macro recall, exact distress, question triggers, and waving responses; lower is better for silent misses and unnecessary questions.}
\label{tab:perception}
\footnotesize
\begin{tabular*}{\textwidth}{@{\extracolsep{\fill}}lccccccc@{}}
\toprule
Perception path & NTU acc.\ (\%)$^{\dagger}$ & Macro recall & Exact distress & Question trigger & Silent miss & Unnecessary question & Waving response \\
\midrule
Pose-centric hybrid & 89.8 & \textbf{0.71} & \textbf{11/16} & \textbf{12/16} & \textbf{4/16} & 2/8 & \textbf{3/4} \\
X3D-UGT, fine-tuned & 97.7 & 0.25 & 1/16 & 2/16 & 14/16 & \textbf{0/8} & 1/4 \\
X3D-UGT, from scratch & 94.8 & 0.29 & 3/16 & 3/16 & 13/16 & 2/8 & 1/4 \\
\bottomrule
\end{tabular*}
\vspace{2pt}

\raggedright\footnotesize $^{\dagger}$Six-way accuracy on a 1,928-clip NTU RGB+D validation subset. The pose-path value is the ST-GCN++ component alone (its rule and geometry components are training-free); checkpoints were test-selected, so benchmark values are optimistic.
\end{table*}

\subsection{Controller protocol}
We bypassed the camera and recognition modules and injected ROS~2 perception and explicit response events on the same topics the live stack uses, with responses issued through a Wizard-of-Oz teleoperation node. Five trials exercised response after the first prompt, and five exercised silence through both prompts. Three boundary trials tested an early response during \textsc{Ask1}, a response during \textsc{Wait2}, and duplicate-alert suppression inside the cooldown. We logged state timestamps, alert count and payload, terminal state, crashes, and deadlocks. This protocol measures deterministic controller behavior; it is not an end-to-end autonomy test.

\section{Results}
Table~\ref{tab:perception} shows a large gap between the perception paths under the robot-camera viewpoint, and benchmark accuracy did not predict robot behavior: between the two directly comparable RGB variants the ordering inverts---97.7\% versus 94.8\% on NTU, but 0.25 versus 0.29 macro recall on the robot. (The pose path's NTU value covers its ST-GCN++ component alone, so cross-path benchmark comparisons are indicative only.) The pose-centric hybrid produced the exact distress cue in 11/16 clips and retained a question-triggering distress cue in 12/16. In contrast, the fine-tuned and from-scratch X3D-UGT variants yielded a question-triggering cue in only 2/16 and 3/16 clips.

\begin{table}[t]
\centering
\caption{Per-class outcomes on the robot-camera clips (correct clips over total; for normal, clips without a false distress tag).}
\label{tab:perclass}
\footnotesize
\setlength{\tabcolsep}{4.0pt}
\begin{tabular}{lccc}
\toprule
Class & Hybrid & X3D-UGT ft.\ & X3D-UGT scr.\ \\
\midrule
\texttt{waving} & 3/4 & 1/4 & 1/4 \\
\texttt{clutching\_chest} & 1/4 & 0/4 & 0/4 \\
\texttt{possible\_collapse} & 3/4 & 1/4 & 3/4 \\
\texttt{staggering} & 4/4 & 0/4 & 0/4 \\
\texttt{headache} & 3/4 & 0/4 & 0/4 \\
\texttt{normal} (no false tag) & 6/8 & 8/8 & 6/8 \\
\bottomrule
\end{tabular}
\end{table}

\begin{figure*}[t]
    \centering
    \includegraphics[width=0.98\textwidth]{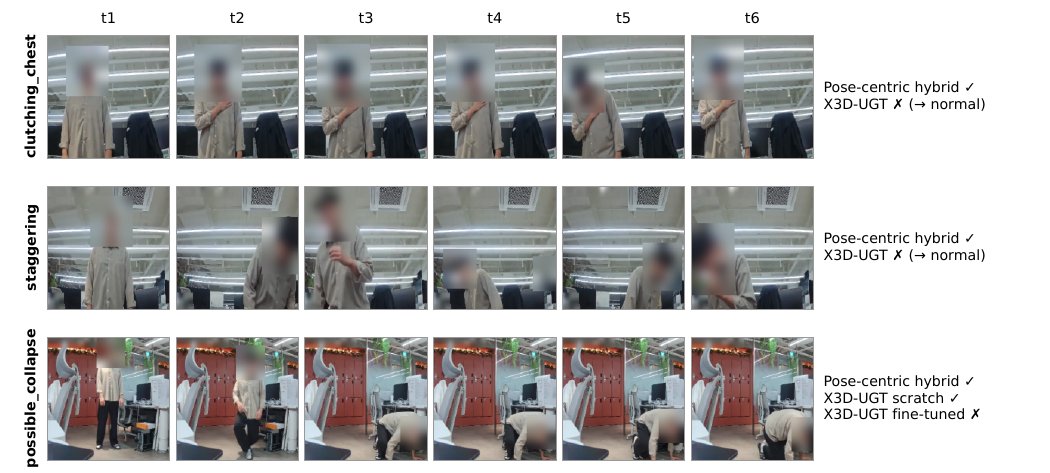}
    \caption{Robot-camera clips (person-crop input, faces blurred) in which the pose-centric hybrid confirmed the distress cue. \texttt{clutching\_chest} and \texttt{staggering} (top, middle) were silently missed by both X3D-UGT variants; \texttt{possible\_collapse} (bottom) was retained by the from-scratch variant only.}
    \label{fig:qualitative}
\end{figure*}

Table~\ref{tab:perclass} and Fig.~\ref{fig:qualitative} localize the gap. Gross-motion cues survived transfer best: the hybrid confirmed \texttt{staggering} in 4/4 clips, and \texttt{possible\_collapse} was the only distress cue the RGB paths retained at all (3/4 for the from-scratch variant). Subtle upper-body cues were hardest: both RGB variants scored 0/4 on \texttt{clutching\_chest}, \texttt{headache}, and \texttt{staggering}, and even the hybrid confirmed \texttt{clutching\_chest} in only 1/4 clips. A person-crop ablation makes a crop-only explanation for the RGB collapse unlikely: across full-frame and four person-crop variants, RGB macro recall stayed between 0.17 and 0.33, and inspection of the cropped inputs shows clearly framed cues that the models still labeled \texttt{normal}.

The interaction-aware measures then localize the consequences. All three hybrid \texttt{clutching\_chest} misses were silent---no tag of any kind was confirmed---and the missed \texttt{headache} clip was confirmed only as hand gestures, so the controller would have had nothing to act on in 4/16 distress clips. Confusions between \texttt{staggering} and \texttt{possible\_collapse} occurred in several clips but are benign at the policy level, because either label opens the same question; one collapse clip triggered the question through a wrong distress label. Both unnecessary questions on normal clips came from spurious \texttt{staggering} tags. Still, the hybrid leaves 4/16 distress clips without any question and would ask unnecessarily in 2/8 normal clips. The small sample does not establish that fine-tuning caused the observed ordering between the two RGB variants, and because the pose path combines keypoints, rules, ST-GCN++, and geometry, its gain cannot be attributed to ST-GCN++ alone.

\begin{table}[t]
\centering
\caption{Event-injection validation of the controller.}
\label{tab:fsm}
\small
\setlength{\tabcolsep}{4.0pt}
\begin{tabular}{lccc}
\toprule
Scenario & Trials & Correct & Alerts \\
\midrule
Response after first prompt & 5 & 5/5 & 0 \\
Silence after both prompts & 5 & 5/5 & 5 (one/trial) \\
Boundary conditions & 3 & 3/3 & as intended \\
\midrule
Total & 13 & \textbf{13/13} & no duplicates \\
\bottomrule
\end{tabular}
\end{table}

All controller trials passed (Table~\ref{tab:fsm}). The first response window opened 2.62~s after a distress event on average (range 2.57--2.67~s), a valid response produced \textsc{Standdown} within 0.19--0.20~s, and two unanswered prompts generated one alert after 21.65~s on average (21.59--21.67~s), close to the nominal 21~s schedule plus timer discretization. The early response was ignored, the second-window response was accepted, and the duplicate alert was suppressed. We observed no unexpected states, deadlocks, or crashes; alert payloads contained the unresolved-distress label, originating cue, and timestamp.

\section{Discussion and Limitations}
The principal design implication is that action recognition and interaction policy should be evaluated together but not conflated: exact labels matter for interpretation, while any confirmed distress label suffices to begin the same low-cost confirmation. The question-first policy changes a false trigger's immediate consequence from an escalation alert to a prompt, but it does not repair a silent miss, so improving cue sensitivity and testing under occlusion and low light remain necessary; our prior characterization of X3D-UGT under controlled low-light, blur, occlusion, and crop stress \cite{yoon2026x3dugt} provides a template for that next step. The per-class pattern also indicates where sensitivity work should concentrate: gross-motion cues transferred far better than subtle upper-body cues, making chest-related motion the natural first target; domain-specific synthetic data generation \cite{yoon2026diffusion} offers one path to densify training coverage for such cues. Finally, question-first escalation is an acceptance mechanism as much as a safety one: camera-based monitoring of older adults often fails at acceptance rather than accuracy \cite{broadbent2009acceptance}, and asking first keeps the person's answer above the algorithm's inference.

The intended deployment is edge-local. The RGB backbone has been characterized on the Jetson Orin Nano---a 5.3~MiB TensorRT FP16 engine at roughly ten clips per second \cite{yoon2026x3dugt}---but the pose path has not yet been profiled on that hardware, so running the full question-first loop at the bedside without streaming video out of the room remains a design goal rather than a measured result. Integrating the validated controller with on-device perception, physical pan--tilt motion, and a real notification channel is the immediate engineering step; the interaction policy itself does not change with that integration.

The present evidence is deliberately narrow. The perception set contains 28 scripted clips from one actor---a controlled intermediate step between benchmark footage and in-the-wild robot-camera sensing---without confidence intervals or demographic variation, and it does not measure false-prompt rates over long deployments. The controller test injects events in process; it bypasses the camera, network inference, physical pan--tilt execution, notification delivery, and human response. Prompts are on-screen text rather than validated speech dialogue, and responses are explicit events or waiting-state waving cues, not speech understanding. We have not tested whether users understand the pan--tilt motions, accept two prompts, or find the timing appropriate for real care contexts. \system{} is therefore neither a medical device nor a validated safety monitor; a larger multi-participant, end-to-end study must evaluate detection, comprehensibility, privacy, burden, and escalation outcomes jointly.

\section{Conclusion}
In a small robot-camera study, a pose-centric hybrid retained far more question-triggering body cues than two X3D-UGT appearance baselines, yet still produced silent misses and unnecessary prompts. The controller, validated in 13/13 event-injection trials, turned each injected cue into two acknowledgment opportunities before a single escalation. The results support question-first interaction as a systems safeguard around imperfect perception and delineate the end-to-end, user-centered validation still required.
\clearpage
\balance
\bibliographystyle{IEEEtran}
\bibliography{references}

@inproceedings{feilseifer2005sar,
  author    = {David J. Feil-Seifer and Maja J. Matari{\'c}},
  title     = {Defining Socially Assistive Robotics},
  booktitle = {Proceedings of the IEEE International Conference on Rehabilitation Robotics},
  pages     = {465--468},
  year      = {2005},
  doi       = {10.1109/ICORR.2005.1501143}
}

@inproceedings{dragan2013legibility,
  author    = {Anca D. Dragan and Kenton C. T. Lee and Siddhartha S. Srinivasa},
  title     = {Legibility and Predictability of Robot Motion},
  booktitle = {Proceedings of the ACM/IEEE International Conference on Human-Robot Interaction},
  pages     = {301--308},
  year      = {2013},
  doi       = {10.1109/HRI.2013.6483603}
}

@inproceedings{ren2023askhelp,
  author    = {Allen Z. Ren and Anushri Dixit and Alexandra Bodrova and Sumeet Singh and Stephen Tu and Noah Brown and Peng Xu and Leila Takayama and Fei Xia and Jake Varley and Zhenjia Xu and Dorsa Sadigh and Andy Zeng and Anirudha Majumdar},
  title     = {Robots That Ask For Help: Uncertainty Alignment for Large Language Model Planners},
  booktitle = {Proceedings of the Conference on Robot Learning},
  series    = {Proceedings of Machine Learning Research},
  volume    = {229},
  pages     = {661--682},
  publisher = {PMLR},
  year      = {2023},
  url       = {https://proceedings.mlr.press/v229/ren23a.html}
}

@inproceedings{feichtenhofer2020x3d,
  author    = {Christoph Feichtenhofer},
  title     = {{X3D}: Expanding Architectures for Efficient Video Recognition},
  booktitle = {Proceedings of the IEEE/CVF Conference on Computer Vision and Pattern Recognition},
  pages     = {203--213},
  year      = {2020},
  doi       = {10.1109/CVPR42600.2020.00028}
}

@inproceedings{yan2018stgcn,
  author    = {Sijie Yan and Yuanjun Xiong and Dahua Lin},
  title     = {Spatial Temporal Graph Convolutional Networks for Skeleton-Based Action Recognition},
  booktitle = {Proceedings of the AAAI Conference on Artificial Intelligence},
  volume    = {32},
  number    = {1},
  pages     = {7444--7452},
  year      = {2018},
  doi       = {10.1609/AAAI.V32I1.12328}
}

@inproceedings{duan2022pyskl,
  author    = {Haodong Duan and Jiaqi Wang and Kai Chen and Dahua Lin},
  title     = {{PYSKL}: Towards Good Practices for Skeleton Action Recognition},
  booktitle = {Proceedings of the 30th ACM International Conference on Multimedia},
  pages     = {7351--7354},
  year      = {2022},
  doi       = {10.1145/3503161.3548546}
}

@inproceedings{yoon2025sr,
  author    = {Dongsik Yoon and Jongeun Kim},
  title     = {Your Super Resolution Model is not Enough for Tackling Real-World Scenarios},
  booktitle = {Proceedings of the IEEE/CVF International Conference on Computer Vision Workshops},
  pages     = {3123--3129},
  year      = {2025}
}

@misc{yoon2026diffusion,
  author = {Dongsik Yoon and Jongeun Kim},
  title  = {From Prompts to Deployment: Auto-Curated Domain-Specific Dataset Generation via Diffusion Models},
  year   = {2026},
  note   = {arXiv:2601.08095}
}

@misc{yoon2026x3dugt,
  author = {Dongsik Yoon and Jongeun Kim and Dayeon Lee},
  title  = {Resource-Efficient {RGB}-Only Action Recognition for Edge Deployment},
  year   = {2026},
  note   = {arXiv:2602.10818}
}

@misc{yolo11ultralytics,
  author       = {Glenn Jocher and Jing Qiu},
  title        = {Ultralytics {YOLO11}},
  year         = {2024},
  note         = {Version 11.0.0},
  howpublished = {\url{https://github.com/ultralytics/ultralytics}}
}

@article{wang2020fallsurvey,
  author  = {Xueyi Wang and Joshua Ellul and George Azzopardi},
  title   = {Elderly Fall Detection Systems: A Literature Survey},
  journal = {Frontiers in Robotics and AI},
  volume  = {7},
  pages   = {71},
  year    = {2020},
  doi     = {10.3389/frobt.2020.00071}
}

@article{robinson2014role,
  author  = {Hayley Robinson and Bruce MacDonald and Elizabeth Broadbent},
  title   = {The Role of Healthcare Robots for Older People at Home: A Review},
  journal = {International Journal of Social Robotics},
  volume  = {6},
  number  = {4},
  pages   = {575--591},
  year    = {2014},
  doi     = {10.1007/s12369-014-0242-2}
}

@misc{kay2017kinetics,
  author = {Will Kay and Jo{\~a}o Carreira and Karen Simonyan and others},
  title  = {The Kinetics Human Action Video Dataset},
  year   = {2017},
  note   = {arXiv:1705.06950}
}

@misc{chen2024wifi,
  author = {Yunwang Chen and Yaozhong Kang and Ziqi Zhao and Yue Hong and Lingxiao Meng and Max Q.-H. Meng},
  title  = {Collaborative Fall Detection and Response Using {Wi-Fi} Sensing and Mobile Companion Robot},
  year   = {2024},
  note   = {arXiv:2407.12537}
}

@inproceedings{lin2019tsm,
  author    = {Ji Lin and Chuang Gan and Song Han},
  title     = {{TSM}: Temporal Shift Module for Efficient Video Understanding},
  booktitle = {Proceedings of the IEEE/CVF International Conference on Computer Vision},
  pages     = {7083--7093},
  year      = {2019}
}

@inproceedings{lin2014coco,
  author    = {Tsung-Yi Lin and Michael Maire and Serge Belongie and James Hays and Pietro Perona and Deva Ramanan and Piotr Doll{\'a}r and C. Lawrence Zitnick},
  title     = {Microsoft {COCO}: Common Objects in Context},
  booktitle = {Proceedings of the European Conference on Computer Vision},
  pages     = {740--755},
  year      = {2014}
}

@inproceedings{lee2023hdgcn,
  author    = {Jungho Lee and Minhyeok Lee and Dogyoon Lee and Sangyoun Lee},
  title     = {Hierarchically Decomposed Graph Convolutional Networks for Skeleton-Based Action Recognition},
  booktitle = {Proceedings of the IEEE/CVF International Conference on Computer Vision},
  year      = {2023}
}

@article{maldonado2019fallen,
  author  = {Saturnino Maldonado-Basc{\'o}n and Cristian Iglesias-Iglesias and Pilar Mart{\'i}n-Mart{\'i}n and Sergio Lafuente-Arroyo},
  title   = {Fallen People Detection Capabilities Using Assistive Robot},
  journal = {Electronics},
  volume  = {8},
  number  = {9},
  pages   = {915},
  year    = {2019},
  doi     = {10.3390/electronics8090915}
}

@misc{azghadi2025fall,
  author = {Seyed Alireza {Rahimi Azghadi} and Truong-Thanh-Hung Nguyen and Helene Fournier and Monica Wachowicz and Rene Richard and Francis Palma and Hung Cao},
  title  = {A Privacy-Preserving Multi-Stage Fall Detection Framework with Semi-supervised Federated Learning and Robotic Vision Confirmation},
  year   = {2025},
  note   = {arXiv:2507.10474}
}

@article{tirziu2025fall,
  author  = {E. T{\^i}rziu and A.-M. Vasilevschi and A. Alexandru and E. Tudora},
  title   = {Real-Time Fall Monitoring for Seniors via {YOLO} and Voice Interaction},
  journal = {Future Internet},
  volume  = {17},
  number  = {8},
  pages   = {324},
  year    = {2025},
  doi     = {10.3390/fi17080324}
}

@article{igual2013fall,
  author  = {Raul Igual and Carlos Medrano and Inmaculada Plaza},
  title   = {Challenges, Issues and Trends in Fall Detection Systems},
  journal = {BioMedical Engineering OnLine},
  volume  = {12},
  number  = {1},
  pages   = {66},
  year    = {2013},
  doi     = {10.1186/1475-925X-12-66}
}

@article{broadbent2009acceptance,
  author  = {Elizabeth Broadbent and Rebecca Stafford and Bruce MacDonald},
  title   = {Acceptance of Healthcare Robots for the Older Population: Review and Future Directions},
  journal = {International Journal of Social Robotics},
  volume  = {1},
  number  = {4},
  pages   = {319--330},
  year    = {2009},
  doi     = {10.1007/s12369-009-0030-6}
}

@inproceedings{shahroudy2016ntu,
  author    = {Amir Shahroudy and Jun Liu and Tian-Tsong Ng and Gang Wang},
  title     = {{NTU RGB+D}: A Large Scale Dataset for 3D Human Activity Analysis},
  booktitle = {Proceedings of the IEEE Conference on Computer Vision and Pattern Recognition},
  pages     = {1010--1019},
  year      = {2016},
  doi       = {10.1109/CVPR.2016.115}
}

@article{liu2020ntu120,
  author  = {Jun Liu and Amir Shahroudy and Mauricio Perez and Gang Wang and Ling-Yu Duan and Alex C. Kot},
  title   = {{NTU RGB+D 120}: A Large-Scale Benchmark for 3D Human Activity Understanding},
  journal = {IEEE Transactions on Pattern Analysis and Machine Intelligence},
  volume  = {42},
  number  = {10},
  pages   = {2684--2701},
  year    = {2020},
  doi     = {10.1109/TPAMI.2019.2916873}
}

\clearpage
\nobalance
\appendices
\section{Nuni Design and Prototype Evolution}
\label{app:design}
This appendix records the design lineage of \system{} from its IEEE RO-MAN 2026 Robot Design Competition proposal to the system evaluated above, so that the question-first controller can be read as one part of an embodied interaction design rather than a bare state machine. Everything introduced here is design specification or development history. Apart from the evaluated-system column of Table~\ref{tab:evolution}, which restates results already reported in Sections~II--IV, none of it was evaluated, and the limitations of Section~V apply unchanged.

\subsection{Prototype hardware}
\begin{figure}[!b]
    \centering
    \includegraphics[width=\columnwidth]{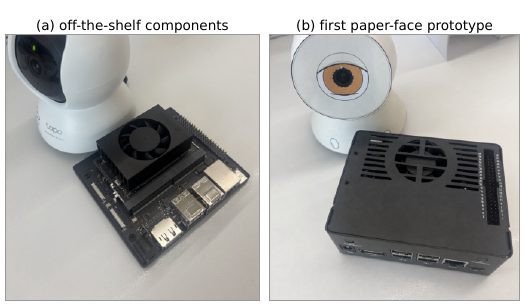}
    \caption{Prototype evolution. (a)~The unmodified TP-Link Tapo pan--tilt camera and the Jetson Orin Nano that serves as the on-device deployment target. (b)~The first paper-face prototype, with the camera lens serving as the pupil.}
    \label{fig:prototype}
\end{figure}

Figure~\ref{fig:prototype} shows the starting point and the first embodied prototype. \system{} is a hacked commodity pan--tilt camera: the gimbal serves as a neck, the lens as a pupil, and a taped paper face as its first skin, so a new face variant costs minutes rather than a fabrication cycle. In the intended bedside configuration the camera streams a low-resolution RTSP substream to a Jetson Orin Nano over a local network and accepts pan--tilt commands through its local API, with a small bedside monitor showing status and the text prompts; as stated in Section~II, the results reported here were instead produced offline on a workstation, and no pan--tilt command was issued in the reported trials. Two properties of this hardware were folded into the design rather than fought: the deliberately paced motor matches the calm tempo intended for a night-time bedside object, and the camera's privacy shutter is mechanical.

\subsection{Physical privacy: not watching as an inspectable state}
The privacy mechanism is physical rather than a software toggle. In the \emph{look away} behavior the head pans to the wall and the lens retracts---an ``eyelid''---so that whether the robot is watching is intended to be legible without trusting a software indicator; whether people read it that way is untested (Section~V). The controller reserves a \textsc{Privacy} state for this behavior, entered on an explicit palm-showing gesture and left on a timeout. It is orthogonal to the escalation path of Fig.~\ref{fig:system} and was not exercised in the reported trials.

\subsection{Behavioral vocabulary}
Table~\ref{tab:vocab} lists the pan--tilt--eyelid recipes specified for each communicative state and the controller state each one serves. The vocabulary is deliberately small and legible; because the gimbal has no roll axis, a quick pan double-take stands in for a head tilt. Prompts in the evaluated build are on-screen text, so the spoken delivery described in the proposal is design intent rather than tested behavior. Because the reported trials scored state outputs only, this vocabulary is the untested half of the loop: the person's answer is read by the robot, but whether the robot's state is read by the person is what the user study of Section~V must establish.

\begin{table}[!b]
\centering
\caption{Behavioral vocabulary (design specification) and the controller state each behavior serves. These motions were not evaluated in this paper. $^{\dagger}$\textsc{Privacy} is orthogonal to the escalation path and is not shown in Fig.~\ref{fig:system}.}
\label{tab:vocab}
\footnotesize
\setlength{\tabcolsep}{3.0pt}
\begin{tabular}{@{}>{\raggedright\arraybackslash}p{0.29\columnwidth}p{0.44\columnwidth}>{\raggedright\arraybackslash}p{0.18\columnwidth}@{}}
\toprule
Behavior & Recipe (pan $\cdot$ tilt $\cdot$ eyelid) & State \\
\midrule
Ambient monitoring, greet & toward the room, lid two-thirds open, near-still; slow pan sweep and a nod on detecting a person & \textsc{Idle} \\
Attending & turn to the person; lean-in tilt; lid fully open & \textsc{Attend} \\
Hesitation, then ask & brief pan away-and-back, a pause, then a forward tilt & \textsc{Ask1}, \textsc{Ask2} \\
Waiting for an answer & hold the attending pose, near-still, so the pause reads as waiting & \textsc{Wait1}, \textsc{Wait2} \\
Stand-down & single nod; lid to half; slow return to home pose & \textsc{Standdown} \\
Reassurance & steady gaze; slow rhythmic nods; occasional soft blink & \textsc{Stay} \\
Privacy & pan to the wall; lid fully shut & \textsc{Privacy}$^{\dagger}$ \\
\bottomrule
\end{tabular}
\end{table}

\subsection{From design concept to the evaluated controller}
Table~\ref{tab:evolution} traces how each element of the proposal's ``ask, listen and watch, escalate only on silence'' concept became the controller of Section~II-B. The direction of every change was toward things that could be measured: a single health-event score became a six-way cue vocabulary with per-class persistence; a spoken question with speech recognition became an on-screen prompt with explicit response events, because speech understanding could not be validated in the available time; and the proposal's roughly 20~s silence timer became two explicitly bounded 8~s response windows, each preceded by a 2.5~s prompt (nominal 21~s), so that a false trigger has two timed chances to be cancelled before one alert. The prompt strings themselves are short by design: \emph{``Are you okay?''}, \emph{``Can you hear me?''}, and, on stand-down, \emph{``Okay. Rest well.''}

\begin{table*}[t]
\centering
\caption{Evolution from the design proposal to the evaluated system.}
\label{tab:evolution}
\footnotesize
\setlength{\tabcolsep}{6.0pt}
\begin{tabular}{@{}p{0.11\textwidth}p{0.37\textwidth}p{0.45\textwidth}@{}}
\toprule
Element & Proposal (mid-2026) & Evaluated system (this paper) \\
\midrule
Trigger & ``a fall or unusual stillness''; one health-event score with a persistence rule & six-way cue vocabulary; per-class persistence (5 consecutive confirmations for chest/head/staggering, 2 for collapse and gestures); static lying excluded by the transition rule \\
Prompt channel & spoken question via offline TTS; listen for an answer & on-screen text prompt; explicit response events, or waving inside a wait window; speech input removed from the evaluated build \\
Escalation & ask, ask once more, alert after roughly 20~s of silence & two prompts with 8~s windows; nominal 21~s; one alert; 15~s duplicate cooldown \\
After the alert & ``I've told them. I'll stay with you.'' & \textsc{Stay} state (designed to nod periodically); alert payload carries label, cue, and timestamp \\
Perception & single lying-on-floor detector planned & RGB appearance vs.\ pose-centric hybrid compared on robot-camera clips \\
Evidence & nightly false-intervention logs planned & 28-clip perception evaluation; 13 event-injection trials; user testing still pending \\
\bottomrule
\end{tabular}
\end{table*}

\subsection{Illustrative scenario}
The proposal framed the policy through a fictional persona: an older adult living alone who had twice refused a monitoring camera. In the scene that motivated the controller, she stumbles on the way back from the bathroom at 3~a.m.; \system{} turns, tilts down, and asks whether she is okay. In the intended common case the scene ends with her answer and a nod (\textsc{Standdown}). If there is no answer, the robot asks once more, waits, sends one alert, and turns back to her to stay (\textsc{Stay}). The scene is illustrative only: the persona is invented, not drawn from interviews or fieldwork. Whether real users find this behavior acceptable remains to be established by the user study called for in Section~V.

\end{document}